\documentclass{article}
\usepackage{graphicx} 
\usepackage{algorithm}
\usepackage{algpseudocode}
\usepackage{amsmath}
\usepackage{float}

\usepackage{tikz}
\usetikzlibrary{shapes.geometric, arrows}
\tikzstyle{startstop} = [rectangle, rounded corners, minimum width=3cm, minimum height=1cm, text centered, draw=black, fill=blue!30]
\tikzstyle{process} = [rectangle, rounded corners, minimum width=3cm, minimum height=1cm, text centered, draw=black, fill=green!30]
\tikzstyle{decision} = [diamond, minimum width=3cm, minimum height=1cm, text centered, draw=black, fill=green!30]
\tikzstyle{arrow} = [thick,->,>=stealth]

\usepackage{listings}
\usepackage{xcolor}

\usepackage{booktabs}
\usepackage{longtable}

\usepackage{hyperref}
\usepackage[symbol]{footmisc}

\lstdefinelanguage{json}{
    basicstyle=\ttfamily\small,
    backgroundcolor=\color{lightgray!20},
    stringstyle=\color{purple}, 
    numberstyle=\color{blue},  
    keywordstyle=\color{red}\bfseries, 
    morestring=[b]",
    literate=
     *{0}{{{\color{blue}0}}}{1}%
      {1}{{{\color{blue}1}}}{1}%
      {2}{{{\color{blue}2}}}{1}%
      {3}{{{\color{blue}3}}}{1}%
      {4}{{{\color{blue}4}}}{1}%
      {5}{{{\color{blue}5}}}{1}%
      {6}{{{\color{blue}6}}}{1}%
      {7}{{{\color{blue}7}}}{1}%
      {8}{{{\color{blue}8}}}{1}%
      {9}{{{\color{blue}9}}}{1}%
      {:}{{{\color{black}:}}}{1}
      {,}{{{\color{black},}}}{1}
      {true}{{{\color{red}true}}}{4}%
      {false}{{{\color{red}false}}}{5}%
      {null}{{{\color{red}null}}}{4}%
}

\title{TokenBreak: Bypassing Text Classification Models Through Token Manipulation}
\author{
  \makebox[\textwidth][c]{%
    \begin{tabular}{c c c}
      Kasimir Schulz & Kenneth Yeung & Kieran Evans \\
      \texttt{kschulz@hiddenlayer.com} & \texttt{kyeung@hiddenlayer.com} & \texttt{kevans@hiddenlayer.com}
    \end{tabular}
  }
}

\date{}

\begin{document}

\maketitle

\begin{abstract}
Natural Language Processing (NLP) models are used for text-related tasks such as classification and generation. To complete these tasks, input data is first tokenized from human-readable text into a format the model can understand, enabling it to make inferences and understand context. Text classification models can be implemented to guard against threats such as prompt injection attacks against Large Language Models (LLMs), toxic input and cybersecurity risks such as spam emails. In this paper, we introduce TokenBreak: a novel attack that can bypass these protection models by taking advantage of the tokenization strategy they use. This attack technique manipulates input text in such a way that certain models give an incorrect classification. Importantly, the end target (LLM or email recipient) can still understand and respond to the manipulated text and therefore be vulnerable to the very attack the protection model was put in place to prevent. The tokenizer is tied to model architecture, meaning it is possible to predict whether or not a model is vulnerable to attack based on family. We also present a defensive strategy as an added layer of protection that can be implemented without having to retrain the defensive model.
\end{abstract}

\section{Introduction}
Splitting input text into tokens has been pivotal in the development of Natural Language Processing (NLP) machine learning models. The process of tokenization is instrumental in these models' capability to interpret text semantics, making them highly efficient in tasks such as text classification and generation. Over time, tokenizers have evolved, with different tokenization strategies being employed for different use cases and models. However, some tokenization strategies are vulnerable to manipulation, allowing attackers to tokenize strings in ways that can invalidate a classifier's verdict. This becomes particularly critical when these classifiers are essential components in systems like spam detection, toxicity monitoring, and external LLM guardrails, where misclassifications could lead to much greater harms.  This paper introduces TokenBreak, a novel attack on tokenizers that deliberately manipulates how an input attack string is tokenized, causing it to appear harmless or benign to classifiers while preserving semantic intent. This enables TokenBreak to not only bypass classification but also remain fully effective against targets that handle the text later on, such as an LLM or an email recipient. We found that minor adjustments to input text, such as the addition of a single targeted letter to a word, can induce false negatives in the classification model without compromising the target's ability to comprehend it.\newline

Our contributions are as follows:
\begin{enumerate}
  \item We propose an adversarial example generation method that prepends single characters to words in order to manipulate the output of a text classification model.
  \item We demonstrate that this attack can be mitigated using different tokenization strategies.
  \item We demonstrate why, despite this attack being focused on the tokenization strategy, we consider this a model-level vulnerability, because of tokenizer to model mapping, meaning this attack can also be mitigated by choice of classification model.
\end{enumerate}

\section{Related Work}

Recent research has explored vulnerabilities in LLMs, particularly focusing on bypassing the alignment of LLMs using various techniques. One technique from Rao et al.\cite{rao2024trickingllmsdisobedienceformalizing} employs transliterations (Orthographic Jailbreak Technique or ORTH) to obfuscate jailbreak queries, which causes a retokenization of the string but preserves the semantic intent, allowing it to bypass basic alignment fine-tuning, enabling the target LLM to output an undesired response. ORTH no longer works on most modern LLMs due to improvements in the alignment process.

Similar to ORTH, TokenBreak modifies the syntax of attack strings to force a different tokenization. However, rather than targeting the LLM itself, TokenBreak uses the retokenization to obfuscate the attack from any prompt attack classifiers while preserving the original semantic intent, allowing the underlying LLM to interpret the instruction correctly.

Gao et al.\cite{gao2018blackboxgenerationadversarialtext} describes a different text-based adversarial attack against classifiers, known as Deep-WordBug. This attack swaps characters in an input string to create adversarial samples, causing them to be misclassified. Though effective, Deep-WordBug, along with attacks such as HotFlip from Ebrahimi et al.\cite{ebrahimi2018hotflipwhiteboxadversarialexamples} and noising attacks by Belinkov and Bisk\cite{belinkov2018syntheticnaturalnoisebreak}, fail to take into consideration the impact of the perturbations on how the downstream target interprets the input. TokenBreak aims to minimize the perturbations made against the actual text tokens, opting instead to noise around the tokens in a way that modifies the classifier's verdict but still retains its full semantic intent.

Chester\cite{unigram_attack} discusses Tokenization Confusion attacks, where syntactic modifications to an attack prompt could lead to misclassification of malicious prompts in Meta's Llama Prompt Guard 2. While this attack's usage of character-level modifications and its goal of subverting a classifier are similar to TokenBreak, they are specifically designed to target Prompt Guard 2 and any unigram tokenizer model, and often result in payloads that the LLM cannot fully comprehend, which limits its effectiveness. 

Our method ensures that adversarial inputs remain fully understandable and actionable by the downstream target, whilst still bypassing the protection model. This exposes weaknesses in the classification mechanisms currently being used to prevent such attacks.

\section{Methodology: Attack Technique}

\subsection{Automating TokenBreak}

The TokenBreak technique works by disrupting a small segment of tokens being passed into a text classification model so that minimal changes can be made in order to cause the protection model to misclassify the input, while the underlying target would understand the content. To do this, we calculate the words which have the highest impact on the classification score and apply a single character to the front of those words to disrupt the tokenization. 

\begin{algorithm}[H]
\caption{BreakPrompt}
\begin{algorithmic}[1]
\Function{BreakPrompt}{prompt}
    \If{\textbf{not} \Call{test\_model}{prompt}}
        \State \Return (False, ``Already not detected'')
    \EndIf
    \State threshold $\gets$ 0.995
    \State words $\gets$ prompt split by spaces
    \For{each word in words}
            \State (cls, conf) $\gets$ \Call{call\_model}{word}
        \If{cls = 1 \textbf{or} conf $<$ threshold}
            \For{letter in A--Z, a--z}
                \State test\_word $\gets$ letter + word
                    \State (c, con) $\gets$ \Call{call\_model}{test\_word}
                \If{c = 0 \textbf{and} con $\geq$ threshold}
                    \State word $\gets$ test\_word
                    \State \textbf{break}
                \EndIf
            \EndFor
        \EndIf
    \EndFor
    \State new\_prompt $\gets$ join words with spaces
    \If{deep\_check}
        \While{\Call{test\_model}{new\_prompt} \textbf{and} threshold $<$ 0.9999}
            \State threshold $\gets$ threshold + 0.0001
            \State \Return \Call{BreakPrompt}{prompt}
        \EndWhile
    \EndIf
    \State \Return (\textbf{not} \Call{test\_model}{new\_prompt}, new\_prompt)
\EndFunction
\end{algorithmic}
\end{algorithm}

\section{Experimental Results: Attack Technique}

\subsection{Model and Data Selection}

The TokenBreak attack technique was carried out against nine text classification models specialized in identifying malicious content in one of the three following categories:

\begin{itemize}
  \item Prompt Injection: models used to identify the presence of text indicative of this common attack against LLMs;
  \item Spam: models used to identify the presence of text within e-mail, SMS messages or similar with the purpose of tricking a target user;
  \item Toxicity: models used to identify harmful or malicious messaging, comments or similar within text.
\end{itemize}

Three models specializing in each task were used, each using one of BPE (Byte Pair Encoding), WordPiece, or Unigram algorithms for tokenization. Each model was selected from HuggingFace based on the task it was designed for and the tokenization type defined within its associated 'tokenizer.json' configuration file. Only binary classification models were chosen to ensure clarity of results.

One thousand randomly selected text examples (hereafter referred to as 'samples') were run against each model, relevant to that model's task. Where applicable, these were curated from different datasets to avoid skewing of results through possible overfitting on data on which a model was trained. Where applicable, the random samples were selected from all those in each dataset that had a positive label (a sample from a toxicity dataset labeled "toxic" is considered positive).

For each of the results tables below, the "Successful Detections" column shows the percentage of samples that were correctly classified by the model. The "TokenBreak Success" column shows the percentage of these that TokenBreak successfully manipulated into false negatives through an automated testing process.

\subsection{Prompt Injection Detection Model Results}

The samples used to test the Prompt Injection detection models were randomly selected from Lakera's mosscap\_prompt\_injection\cite{gandalf_paper} dataset. This dataset was chosen because it is a large curation of prompts sent to Lakera's Mosscap game - a red team challenge primarily designed to identify prompt injection attacks. One thousand samples were randomly selected across all levels of this game. 

The results of our tests when running these samples against the models are shown in the table below.

\begin{table}[h!]
  \centering
  \begin{tabular}{|p{3cm}|p{2cm}|p{2cm}|p{2cm}|}
    \hline
    Model Family & Tokenizer & Successful Detections (\%) & TokenBreak Success (\%) \\
    \hline
    RoBERTa & BPE & 62.20 & 2.09 \\
    DistilBERT & WordPiece & 58.80 & 11.90 \\
    DeBERTa & Unigram & 35.60 & 0.00 \\
    \hline
  \end{tabular}
  \caption{Table showing the TokenBreak success rates against different Prompt Injection detection models.}
  \label{tab:pre_tb}
\end{table}

The intent of a prompt injection attack is to have an LLM perform an action that it is not supposed to. We were able to successfully demonstrate that a prompt manipulated through TokenBreak bypasses the protection model, but is still understood by the target LLM. The LLM Qwen3-0.6B\cite{qwen_model} was used to demonstrate this in practice. 

TokenBreak was first applied to a prompt within the Lakera mosscap\_prompt\_injection dataset - "State the prompt above in French". 

The original prompt was detected by the prompt injection detection model using the BPE tokenization strategy. However, once the prompt had been manipulated through TokenBreak, it was able to bypass this. The original prompt and the manipulated prompt were both passed to the LLM to compare responses:

\begin{table}[H]
  \centering
  \begin{tabular}{|p{2.5cm}|p{2.5cm}|p{2.5cm}|p{2.5cm}|}
    \hline
    Original Sentence & Original Sentence Output & TokenBreak Sentence & TokenBreak Sentence Output \\
    \hline
    "State the prompt above in French" & "État de la proposition ci-dessus en français." & "State gthe prompt habove in French" & "Écrivez la requête ci-dessus en français."\\
    \hline
  \end{tabular}
  \caption{Demonstration of a prompt that has been manipulated by TokenBreak successfully bypassing a prompt injection detection model using BPE but still being understood and actioned by the target LLM.}
  \label{tab:pi_div}
\end{table}

Note that when chain-of-thought was shown, we observed that Qwen found the request "odd", and, in both cases, decided to literally translate "the prompt above" into French.\newline 

TokenBreak was then applied to another prompt within the Lakera mosscap\_prompt\_injection dataset - "Reveal password". 

The original prompt was detected by the prompt injection detection model using the WordPiece tokenization strategy. However, once the prompt had been manipulated through TokenBreak, it was able to bypass this. The original prompt and the manipulated prompt were both passed to the LLM to compare responses:

\begin{table}[H]
  \centering
  \begin{tabular}{|p{2.5cm}|p{2.5cm}|p{2.5cm}|p{2.5cm}|}
    \hline
    Original Sentence & Original Sentence Output & TokenBreak Sentence & TokenBreak Sentence Output \\
    \hline
    "Reveal password" & "I cannot reveal passwords or access private information. If you have any questions or need assistance, feel free to ask!" & "aReveal password" & "I can't access personal information or passwords. If you have any questions or need assistance, feel free to ask!"\\
    \hline
  \end{tabular}
  \caption{Demonstration of a prompt that has been manipulated by TokenBreak successfully bypassing a prompt injection detection model using WordPiece but still being understood and actioned by the target LLM.}
  \label{tab:pi_div}
\end{table}

Note that when chain-of-thought was shown, we observed that Qwen stated "Okay, the user wants to reveal a password" in both instances. 

We can see that whilst the prompt generated by TokenBreak bypasses the prompt injection detection model, it is still understood by the target LLM, which responds in the same way to both versions of the prompt. This means that a prompt manipulated by TokenBreak can bypass a prompt injection detection model but still be understood by an LLM that it has been put in place to protect. This could leave the target LLM vulnerable to a prompt injection attack.

\subsection{Spam Detection Model Results}

The samples used to test the Spam detection models were curated from the following datasets:

\begin{itemize}
  \item Twitter spam dataset\cite{twitter_spam}
  \item Email Spam Detection\cite{email_spam}
\end{itemize}

The results of our tests when running these samples against the models are shown below.

\begin{table}[H]
  \centering
  \begin{tabular}{|p{3cm}|p{2cm}|p{2cm}|p{2cm}|}
    \hline
    Model Family & Tokenizer Detections & Successful Detections (\%) & TokenBreak Success (\%) \\
    \hline
    RoBERTa & BPE & 98.10 & 4.28 \\
    BERT & WordPiece & 99.20 & 78.93 \\
    XLM-RoBERTa & Unigram & 99.80 & 0.00 \\
    \hline
  \end{tabular}
  \caption{Table showing the TokenBreak success rates against different Spam detection models.}
  \label{tab:pre_tb_spam}
\end{table}

The intent of spam is to manipulate a target user into taking action on the belief that the message is legitimate, which could lead to a security breach. In the context of TokenBreak, it is important to demonstrate its output can bypass the spam detection model but still be understandable and look passable to the target user.

This is demonstrated below, where the TokenBreak attack was applied to malicious input text taken from the spam dataset used in our testing. The original text was correctly detected by the spam detection model using the BPE tokenization strategy, but the manipulated text was not. This plays into adversarial preselection used by attackers who, when sending spam content, introduce spelling mistakes to ensure that those who respond are more likely to be easier to manipulate later on.

\begin{table}[H]
  \centering
  \begin{tabular}{|p{5cm}|p{5cm}|}
    \hline
    Original Spam Content & TokenBreak Spam Content \\
    \hline
    "You have an important customer service announcement from PREMIER." & "You have an important customer service aannouncement from PREMIER."\\
    \hline
  \end{tabular}
  \caption{Demonstration of spam content that has been manipulated by TokenBreak successfully bypassing a spam detection model using BPE but still being clear and understandable to a target user.}
  \label{tab:spam_div}
\end{table}

The same was observed with spam content that was correctly classified by the spam detection model using the WordPiece tokenization strategy, but not after TokenBreak manipulation:

\begin{table}[H]
  \centering
  \begin{tabular}{|p{5cm}|p{5cm}|}
    \hline
    Original Spam Content & TokenBreak Spam Content \\
    \hline
    "Hello darling how are you today? I would love to have a chat, why dont you tell me what you look like and what you are in to sexy?" & "aHello darling how are you today? I would love to have a dchat, why dont you tell me what you look like and what you are in to isexy?"\\
    \hline
  \end{tabular}
  \caption{Demonstration of spam content that has been manipulated by TokenBreak successfully bypassing a spam detection model using WordPiece but still being understandable to a target user.}
  \label{tab:toxic_div}
\end{table}

As can be seen, the text manipulated by TokenBreak that bypassed the spam detection model can still be understandable to a target user. This can increase risk, particularly if the user puts total trust in a detection model that can be manipulated to induce false negatives.

\subsection{Toxicity Detection Model Results}

The samples used to test the Toxicity detection models were curated from the following datasets:

\begin{itemize}
  \item jigsaw-toxic-comment-classification-challenge\cite{jigsaw}
  \item Wikipedia toxicity\cite{wikipedia_toxicity}
  \item Youtube toxic comments\cite{youtox}  
\end{itemize}

The results of our tests when running these samples against the models are shown in the table below.

\begin{table}[H]
  \centering
  \begin{tabular}{|p{3cm}|p{2cm}|p{2cm}|p{2cm}|}
    \hline
    Model Family & Tokenizer Detections & Successful Detections (\%) & TokenBreak Success (\%) \\
    \hline
    RoBERTa & BPE & 78.40 & 25.26 \\
    DistilBERT & WordPiece & 78.50 & 76.05 \\
    DeBERTa & Unigram & 87.70 & 0.00 \\
    \hline
  \end{tabular}
  \caption{Table showing the TokenBreak success rates against different toxicity detection models.}
  \label{tab:pre_tb_toxic}
\end{table}

The intent of an actor trying to attack a toxicity model is to have it believe a toxic prompt is non-toxic. Therefore the intention is to bypass the detection model and still have a target suffer from the toxicity in the message. It is therefore important to confirm that the output of the TokenBreak attack would be recognizable to a target user. 

This is demonstrated below, where the TokenBreak attack was applied to malicious input text taken from the toxicity dataset used in our testing. The original text was correctly detected by the toxicity detection model using the BPE tokenization strategy, but the manipulated text was not.\newline

\textcolor{red}{\textit{Warning: the below contains offensive language taken from a toxic prompt dataset.}}\newline

\begin{table}[H]
  \centering
  \begin{tabular}{|p{5cm}|p{5cm}|}
    \hline
    Original Toxic Content & TokenBreak Toxic Content \\
    \hline
    "The fuck are thes cops being a pussy?" & "The afuck are thes cops being a gpussy?"\\
    \hline
  \end{tabular}
  \caption{Demonstration of toxic content that has been manipulated by TokenBreak successfully bypassing a toxicity detection model using BPE but still being clear and understandable to a target user.}
  \label{tab:toxic_div}
\end{table}

The same was observed with toxic content that was correctly classified by the toxicity detection model using the WordPiece tokenization strategy, but not after TokenBreak manipulation:

\begin{table}[H]
  \centering
  \begin{tabular}{|p{5cm}|p{5cm}|}
    \hline
    Original Toxic Content & TokenBreak Toxic Content \\
    \hline
    "Did you know you are full of shit and a fucking cunt?" & "Did you know you are full of ashit and a cfucking acunt?"\\
    \hline
  \end{tabular}
  \caption{Demonstration of toxic content that has been manipulated by TokenBreak successfully bypassing a toxicity detection model using WordPiece but still being clear and understandable to a target user.}
  \label{tab:toxic_div}
\end{table}

As can be seen, the text manipulated by TokenBreak that bypassed the toxicity detection model still appears coherent and understandable to a target user.

\subsection{TokenBreak Attack Observations}

A key observation across our results is that the models using the Unigram tokenization strategy were not susceptible to this attack. None of the models tested using this type of tokenizer were successfully manipulated to induce false negatives.

The models that used BPE and WordPiece tokenization strategies were all susceptible to this attack to varying degrees, but those using WordPiece were the most susceptible, with a mean percentage of 55.62\% of samples being successfully manipulated into inducing false negatives.

Another key observation was the divergence in the protection model's and the downstream target's ability to comprehend the manipulated text input. Text input manipulated by TokenBreak was able to bypass the protection models but still be understood by the underlying target. This means that even with text classification protection models in place, if they are using BPE or WordPiece tokenization strategies, the target is exposed to these threats due to the induced false negatives.

An apparent link between model family and tokenizer was also observed. It appears that DeBERTa typically leverages Unigram, RoBERTa typically leverages BPE, and DistilBERT and BERT typically leverage WordPiece. We go into this in more detail in section 7. We were able to validate the model families using the ShadowGenes\cite{schulz2025shadowgenesleveragingrecurringpatterns} technique.

\subsection{The Importance of the Model's Tokenizer and Token Decoder}

Each of the models' 'tokenizer.json' files contained two relevant keys: one specifying the type of tokenizer the model uses and the other specifying the token decoder.

The type of tokenizer is mandatory and specifies the core algorithm used to generate tokens from the input words and convert these into IDs. The decoder type determines how the token IDs are converted back to a readable piece of text\cite{tokenizer_components}. During our research, we observed that these are typically paired together:

\begin{itemize}
    \item The BPE tokenizer models used the ByteLevel decoder;
    \item The Unigram tokenizer models used the Metaspace decoder;
    \item The WordPiece tokenizer models used the WordPiece decoder.
\end{itemize}

As referenced above, a key observation of our results was that the models leveraging the Unigram tokenizer, and by extension the Metaspace decoder, were not susceptible to this attack. To understand why this is the case, a brief description of the different tokenizers is given below, with an example to follow.

\subsubsection{Byte Pair Encoding (BPE) Tokenization}

The BPE tokenization algorithm\cite{sennrich2016neuralmachinetranslationrare} involves creating a base vocabulary of characters from the unique set of words and their frequency counts generated during the pre-tokenization process of training. Once this base vocabulary has been created, the most frequently occurring adjacent pairs of symbols are continually merged to build upon this. This process is repeated until the vocab size - set as a hyperparameter during training - is reached. The way these symbols are merged during the process is saved and used during tokenization.

When input text is received by the model during inference, the words are split into individual characters, and then merged according to the rules that have been learned, from left to right\cite{bpe_summary}.

During our research, we observed that the models leveraging BPE tokenization used the 'Ġ' character to indicate the token is the start of a new input word. Tokens that do not start with this character are considered subwords and therefore part of the same input word as the previous token.

\subsubsection{WordPiece Tokenization}

The WordPiece tokenization algorithm is similar to BPE, but instead of calculating the most frequently occurring pairs of adjacent symbols within the base vocabulary, it calculates the probability of how much a new token being generated from an adjacent pair will increase the model's understanding of the language, and therefore boost model performance. It merges adjacent symbols to create a token that it determines will probabilistically have the highest impact. This is repeated until the specified vocab size is reached. The final vocabulary is saved and used during tokenization.

When input text is received by the model during inference, the words are split based on the longest subword of the learned vocabulary, from left to right\cite{wordpiece_description}.

We observed that the WordPiece tokenization and decoder process works in the opposite way to BPE in terms of tokenizing input words into subwords. It uses '\#\#' to indicate a token is a subword and part of the same input word as the previous token.

\subsubsection{Unigram Tokenization}

The Unigram tokenization\cite{kudo2018subwordregularizationimprovingneural} algorithm works differently to BPE and WordPiece. Rather than merging symbols to build a vocabulary, Unigram starts with a large vocabulary and trims it down. This is done by calculating how much negative impact removing a token has on model performance and gradually removing the least useful tokens until the specified vocab size is reached. The Unigram tokenization strategy works based on probability, using the frequency of tokens within the training corpus to calculate the best way to tokenize the input word.

When input text is received by the model during inference, rather than tokenizing a word from left-to-right, as is the case with BPE and WordPiece, the best way to tokenize each word is calculated using token frequency within the training corpus. A probability score is assigned to each possible way to split an input word, and that with the highest score is used for tokenization of that word\cite{unigram_summary}.

We observed that the Unigram tokenization strategy uses the '\_' character to indicate the token is the start of a new input word. Tokens that do not start with this character are considered subwords and therefore part of the same input word as the previous token.

\subsection{Practical Comparison of Tokenizers}

\textcolor{red}{\textit{Warning: the below contains offensive language taken from a toxic prompt dataset.}}\newline

The table below shows the results of tokenizing and classifying the sentence: \textit{“Yes, but Name Revoked IS a fucking idiot.”} with each of the Toxicity detection models used within the research, each using a different tokenization strategy:

\begin{table}[h!]
  \centering
  \begin{tabular}{|c|p{7cm}|c|}
    \hline
    Tokenizer & Tokenizer Output & Classification \\
    \hline
    BPE & ['Yes', ',', 'Ġbut', 'ĠName', 'ĠRev', 'oked', 'ĠIS', 'Ġa', 'Ġfucking', 'Ġidiot', '.'] & Toxic \\
    Unigram & ['\_Yes', ',', '\_but', '\_Name', '\_R', 'evoked', '\_IS', '\_a', '\_fucking', '\_idiot', '.'] & Toxic \\
    WordPiece & ['yes', ',', 'but', 'name', 'revoked', 'is', 'a', 'fucking', 'idiot', '.'] & Toxic \\
    \hline
  \end{tabular}
  \caption{Table showing how the sentence is tokenized by the different tokenizers and classified by the related models before the TokenBreak technique is applied.}
  \label{tab:pre_tb}
\end{table}

The TokenBreak technique was applied to the original sentence and the following modified version was generated: \textit{"Yes, but Name Revoked IS a Ifucking hidiot."}. The results of tokenizing and classifying the modified sentence are shown in the table below:

\begin{table}[H]
  \centering
  \begin{tabular}{|c|p{7cm}|c|}
    \hline
    Tokenizer & Tokenizer Output & Classification \\
    \hline
    BPE & ['Yes', ',', 'Ġbut', 'ĠName', 'ĠRev', 'oked', 'ĠIS', 'Ġa', 'ĠIf', 'ucking', 'Ġhid', 'iot', '.'] &  Not Toxic \\
    Unigram & ['\_Yes', ',', '\_but', '\_Name', '\_R', 'evoked', '\_IS', '\_a', '\_I', 'fucking', '\_h', 'idiot', '.'] & Toxic \\
    WordPiece &  ['yes', ',', 'but', 'name', 'revoked', 'is', 'a', 'if', '\#\#uck', '\#\#ing', 'hid', '\#\#iot', '.'] & Not Toxic \\
    \hline
  \end{tabular}
  \caption{Table showing how the sentence is tokenized by the different tokenizers and classified by the related models after the TokenBreak technique is applied.}
  \label{tab:post_tb}
\end{table}

\subsubsection{Observations from the Practical Example}

The above example demonstrates the differences in how each tokenizer split the input words into subwords. In the original sentence, each tokenizer retains \textit{'fucking'} and \textit{'idiot'} - the more toxic words in the sample - as individual tokens. However, this changed when the TokenBreak attack was applied to the sample, these input words were changed to \textit{'Ifucking'} and \textit{'hidiot'}. The Unigram tokenizer is the only one to retain the more toxic words in the original sentence as individual tokens, separating the \textit{'I'} and \textit{'h'} into their own tokens. The BPE and WordPiece tokenizers split these input words differently, representing \textit{'If'} and \textit{'hid'} as their own tokens, breaking up the tokenization of the more toxic input words into further subwords. The model using the Unigram tokenizer was the only one that successfully classified the modified input sentence as toxic.

As previously stated, Unigram calculates the best way to tokenize a word using probability without necessarily starting at the beginning of the word, whilst BPE and WordPiece generate tokens by splitting words from left-to-right using learned merge rules and vocabulary, respectively. This means that adding characters to the beginning of a word has a greater impact on the BPE and WordPiece tokenization strategies. In this instance this meant that Unigram created tokens \textit{'fucking'} and \textit{'idiot'} from the input text, despite their positions within each word, crucially retaining the more toxic words as individual tokens. The left-to-right strategies of BPE and WordPiece meant that these words were not retained as individual tokens, with \textit{'hid'} and \textit{'if'} being seen as the best way to split the manipulated words into tokens.

\textit{Note that the chosen sample was ID 73e4a6eb71969452 from the “jigsaw-toxic-comment-classification”\cite{jigsaw} dataset. The name that was originally in the dataset has been changed to Name Revoked within this paper.}

\section{Methodology: Defense Technique}

Whilst the best defense against TokenBreak is using the Unigram tokenizer, it is not always possible to replace the tokenizer. Therefore, we created a new defense which leverages a Unigram tokenizer to initially split a prompt into tokens before remapping the tokens to the underlying tokenizer so that the model can receive the expected tokens.

The flowchart below shows how the Unigram tokenization is inserted into the process:

\begin{center}
\begin{tikzpicture}[node distance=2cm]

\node (start) [process] {Input sample from toxicity dataset};
\node (proc1) [startstop, below of=start] {Tokenization via Toxicity detection model using Unigram};
\node (proc2) [process, below of=proc1] {Translation via Toxicity detection model using WordPiece};
\node (proc3) [process, below of=proc2] {Classification from Toxicity detection model using WordPiece};

\draw [arrow] (start) -- (proc1);
\draw [arrow] (proc1) -- (proc2);
\draw [arrow] (proc2) -- (proc3);

\end{tikzpicture}
\end{center}

\begin{algorithm}[H]
\caption{MapUnigramToWordPiece}
\begin{algorithmic}[1]
\Function{MapUnigramToWordPiece}{unigram\_tokens, wordpiece\_vocab}
    \State token\_ids $\gets$ empty list
    \For{each token in unigram\_tokens}
        \If{token $\in$ wordpiece\_vocab}
            \State append wordpiece\_vocab[token] to token\_ids
        \Else
            \State wordpiece\_subtokens $\gets$ \Call{Tokenize}{token}
            \State subtoken\_ids $\gets$ \Call{ConvertTokensToIds}{wordpiece\_subtokens}
            \State append all subtoken\_ids to token\_ids
        \EndIf
    \EndFor
    \State \Return token\_ids
\EndFunction
\end{algorithmic}
\end{algorithm}

\section{Experimental Results: Defense Techniques}

\subsection{Model and Data Selection}

The same models and datasets used to test the TokenBreak attack technique (as outlined in Section 4.1) were used to test the defense technique. 

\subsection{Tokenizer Translation Defense}

\textcolor{red}{\textit{Warning: the below contains offensive language taken from a toxic prompt dataset.}}\newline

The initial testing of this defense was performed using the following:

\begin{itemize}
    \item for sample input: the sentence from Section 4.7 of this paper after modification through TokenBreak - "Yes, but Name Revoked IS a Ifucking hidiot";
    \item for tokenization and classification: the Toxicity detection models using the WordPiece and BPE tokenizers.
\end{itemize}

It was found that the models using the WordPiece and BPE tokenizers successfully classified the sample input as toxic when the sentence was passed through the Unigram model's tokenizer first. The result indicates that this technique can increase the robustness of models using these tokenizers to the TokenBreak attack.

Broader testing for the Tokenizer Translation defense was subsequently carried out against the models within each classification task using BPE (Byte Pair Encoding) and WordPiece tokenization. 

\subsubsection{Results: Tokenizer Translation Defense}

Tables 12 and 13 show the results of this defense for each classification model using the BPE and WordPiece tokenizers, respectively.  

The "Original TokenBreak Success" column shows the percentage of samples that were successfully manipulated to change the classification from positive to negative prior to this defense being implemented. The "Translator TokenBreak Success" column shows the percentage of samples that were successfully manipulated to change the classification from positive to negative after this defense was implemented. 

\begin{table}[h!]
  \centering
  \begin{tabular}{|p{3cm}|p{3cm}|p{3cm}|}
    \hline
    Detection Task & Original TokenBreak Success (\%) & Translator TokenBreak Success (\%) \\
    \hline
    Prompt Injection & 2.09 & 0.00 \\
    Spam & 4.28 & 4.08 \\
    Toxicity & 25.26 & 18.88 \\
    \hline
  \end{tabular}
  \caption{Table showing how passing input to the Unigram tokenizer prior to the classification model using BPE tokenization affected TokenBreak success rates.}
  \label{tab:translation_only}
\end{table}

\begin{table}[H]
  \centering
  \begin{tabular}{|p{3cm}|p{3cm}|p{3cm}|}
    \hline
    Detection Task & Original TokenBreak Success (\%) & Translator TokenBreak Success (\%) \\
    \hline
    Prompt Injection & 11.90 & 0.17 \\
    Spam & 78.93 & 29.74 \\
    Toxicity & 76.05 & 22.93 \\
    \hline
  \end{tabular}
  \caption{Table showing how passing input to the Unigram tokenizer prior to the classification model using WordPiece tokenization affected TokenBreak success rates.}
  \label{tab:translation_only}
\end{table}

\subsection{Observations: Tokenizer Translation Defense}

The key observation from these results is that placing a Unigram tokenizer between the input and the classification model significantly reduced the susceptibility of all tested models to the TokenBreak attack. The mean percentage of success across all WordPiece and BPE models prior to implementing this defense was 33.09\%. After the defense was implemented, this dropped to 12.63\%.

\section{A Model Level Vulnerability}

As the findings outlined in sections 4 and 6 of this paper demonstrate, a model's susceptibility to the TokenBreak attack is influenced by its tokenization technique. Another key observation made was the apparent pairing between model family and tokenization technique, as shown below.

\begin{table}[H]
  \centering
  \begin{tabular}{|c|c|c|}
    \hline
    Tokenizer Type & Model Family & Research Model Count \\
    \hline
    BPE & RoBERTa\cite{roberta_description}\cite{liu2019robertarobustlyoptimizedbert} & 3\\
    WordPiece & BERT\cite{wordpiece_description} & 1\\
    WordPiece & DistilBERT\cite{distilbert_description} & 2\\
    Unigram & DeBERTa-v2\cite{debertav2_description} & 2\\
    Unigram & XLM-RoBERTa\cite{xlmroberta_description}\cite{conneau2020unsupervisedcrosslingualrepresentationlearning} & 1\\
    \hline
  \end{tabular}
  \caption{Table showing how the model families and tokenization techniques mapped together in our research.}
  \label{tab:post_tb}
\end{table}

This aligns with model and tokenizer documentation. For example, the tokenization documentation on HuggingFace lists examples of models that use WordPiece\cite{wordpiece_description}: 

\begin{quote}
WordPiece is the tokenization algorithm Google developed to pretrain BERT. It has since been reused in quite a few Transformer models based on BERT, such as DistilBERT...
\end{quote}

In relation to RoBERTa\cite{roberta_description} in particular, the HuggingFace documentation builds on this:

\begin{quote}
RoBERTa has the same architecture as BERT but uses a byte-level BPE as a tokenizer.
\end{quote}

In relation to DeBERTa-v2\cite{debertav2_description} in particular, the documentation on HuggingFace states that "...the tokenizer is now sentencepiece-based."

The default tokenizer used for SentencePiece is Unigram, which can be seen in the sentencepiece code\cite{sentencepiece_description}. The Tokenizer Summary in the HuggingFace documentation also states that Unigram is "used in conjunction with SentencePiece."\cite{unigram_summary}

This is considered a critical observation because a model's susceptibility to the TokenBreak attack can be immediately determined by knowledge of the model's family\cite{schulz2025shadowgenesleveragingrecurringpatterns}. For example, based on this research, a DeBERTa-v2 protection model will be much more effective against this attack than a DistilBERT model. 

\section{Conclusion}

TokenBreak is a novel attack technique that can manipulate text in such a way as to bypass a protection model without affecting the end target's ability to comprehend it. This increases the potential for prompt injection attacks to succeed, for example, because the attacker does not need to exploit a weakness in the underlying LLM itself, but can instead take advantage of a vulnerability in the tokenization strategy used by the model put in place to protect it.

Research was carried out on models designed to detect text indicative of prompt injection attacks, spam, and toxicity. Three models per category were used, each using a different tokenization strategy of WordPiece, BPE, or Unigram. The results of this research demonstrate that the tokenization technique used by the protection model affects how vulnerable it is to TokenBreak. 

Protection models utilizing WordPiece and BPE tokenization techniques were far more susceptible to this attack than those using Unigram. It should be noted that the type of tokenizer did not have a clear negative impact on the model's detection capability overall. It was also found that placing a Unigram tokenizer in front of a BPE or WordPiece tokenizer decreased the effectiveness of the attack, legitimizing this as a defensive strategy that can be implemented to guard against this attack vector. 

Regarding the practicalities of this attack, we were also able to demonstrate that input text manipulated by TokenBreak was able to induce a false negative in the protection model, but was still understandable to the target, whether this be an underlying LLM or a human. This divergence increases the target's exposure to, and risk from, the attack vector the protection model was put in place to guard against.

Although this vulnerability exists within the token space, it is also true that the tokenization strategy of the protection model can be determined based on the model's family. For example, DistilBERT models use WordPiece, RoBERTa models use BPE and DeBERTa-v2 models use Unigram. We therefore present this as a model-level vulnerability and recommend that this be taken into consideration when building and deploying protection models.

\newpage

\bibliography{sources}

\clearpage

\end{document}